\documentclass[10pt]{IEEEtran}
\IEEEoverridecommandlockouts
\usepackage{cite}
\usepackage{amsmath,amssymb,amsfonts}
\usepackage{graphicx}
\usepackage{textcomp}
\usepackage{booktabs}
\usepackage[font=small,skip=4pt]{caption}
\usepackage{multirow}
\usepackage{hyperref}
\hypersetup{hidelinks}
\usepackage{url}
\usepackage{nicefrac}
\usepackage{microtype}
\usepackage{xcolor}
\usepackage{algorithm}
\usepackage{algpseudocode}
\usepackage{needspace}
\usepackage{placeins}
\usepackage{tikz}
\usepackage{orcidlink}
\usepackage{comment}
\usepackage{braket}
\usepackage{listings}
\usepackage{subcaption}
\usepackage{tabularx}
\usepackage{array}
\usepackage{makecell}
\usepackage{colortbl}
\usepackage{setspace}
\lstdefinestyle{qpy}{%
  language=Python,
  basicstyle=\ttfamily\scriptsize\color{black},
  keywordstyle=\color{black},
  commentstyle=\color{black}\itshape,
  stringstyle=\color{black},
  identifierstyle=\color{black},
  numbers=left, numberstyle=\tiny\color{black!55}, numbersep=6pt,
  keepspaces=true, tabsize=4, showstringspaces=false,
  breaklines=true, breakatwhitespace=true, breakautoindent=true,
  frame=none,
  xleftmargin=2em,
  aboveskip=6pt, belowskip=2pt, escapeinside={(*@}{@*)}}

\usepackage[most]{tcolorbox}
\newtcolorbox{casebox}[1]{%
  enhanced, breakable, sharp corners, boxrule=0.6pt,
  colframe=gray!55, colback=white,
  colbacktitle=blue!9, coltitle=black,
  fonttitle=\bfseries\footnotesize, halign title=flush left,
  title={#1}, left=5pt, right=5pt, top=3pt, bottom=3pt,
  toptitle=2.5pt, bottomtitle=2.5pt,
  before skip=6pt, after skip=4pt}

\newtcolorbox{codefigurebox}[1]{%
  enhanced, breakable=false, sharp corners, boxrule=0.6pt,
  colframe=gray!55, colback=white,
  colbacktitle=blue!9, coltitle=black,
  fonttitle=\bfseries\footnotesize, halign title=flush left,
  title={#1}, left=5pt, right=5pt, top=3pt, bottom=3pt,
  toptitle=2.5pt, bottomtitle=2.5pt,
  before skip=4pt, after skip=2pt}

\newtcolorbox{codesnippet}[1]{%
  enhanced, breakable=false, sharp corners, boxrule=0.45pt,
  colframe=gray!40, colback=gray!4,
  colbacktitle=gray!12, coltitle=black,
  fonttitle=\bfseries\scriptsize, halign title=flush left,
  title={#1}, left=4pt, right=4pt, top=2pt, bottom=2pt,
  toptitle=2pt, bottomtitle=2pt,
  before skip=4pt, after skip=2pt}

\newtcolorbox{casestudy}[1]{%
  enhanced, breakable=false, sharp corners, boxrule=0.45pt,
  colframe=blue!35, colback=blue!2,
  colbacktitle=blue!12, coltitle=black,
  fonttitle=\bfseries\scriptsize, halign title=flush left,
  title={#1}, left=3pt, right=3pt, top=1pt, bottom=1pt,
  toptitle=1.5pt, bottomtitle=1.5pt,
  before skip=2pt, after skip=1pt}

\newcommand{\casemetarow}[4]{%
  \par\vspace{2pt}{\scriptsize
  \textbf{Metrics:} ES=#1 \quad CS=#2 \quad BDS=#3 \quad TQS=#4\par}}
\newcommand{\casetakeaway}[1]{%
  \par\vspace{1pt}{\scriptsize\textit{Note:} #1\par}}

\newcolumntype{L}[1]{>{\raggedright\arraybackslash}p{#1}}

\newcommand{\tablelegend}[1]{%
  \par\vspace{4pt}%
  \begin{minipage}{\columnwidth}%
  \scriptsize\textit{Legend:} #1%
  \end{minipage}}

\newcommand{\testbad}[1]{\cellcolor{red!16}\textbf{#1}}

\newcommand{\oracle}[1]{\cellcolor{blue!8}#1}

\def\BibTeX{{\rm B\kern-.05em{\sc i\kern-.025em b}\kern-.08em
    T\kern-.1667em\lower.7ex\hbox{E}\kern-.125emX}}

\begin{document}

\title{QTestBench: A Test Framework for Evaluating LLM-Generated Quantum Code}

\author{\IEEEauthorblockN{Meriem Terki\textsuperscript{1}, Nouhaila Innan\textsuperscript{2,3}, Minghao Shao\textsuperscript{2},\\ Muhammad Kashif\textsuperscript{2,3}, Alberto Marchisio\textsuperscript{2,3,4}$^*$\thanks{$^*$This work has been carried out while at NYUAD.},
Muhammad Shafique\textsuperscript{2,3}}\\
\vspace{2mm}
\IEEEauthorblockA{%
\textsuperscript{1}\normalsize  École Supérieure en Informatique de Sidi Bel Abbès, Algeria\\
\normalsize \textsuperscript{2}eBRAIN Lab, Division of Engineering, New York University Abu Dhabi (NYUAD), Abu Dhabi, UAE\\
\normalsize\textsuperscript{3}Center for Quantum and Topological Systems (CQTS), NYUAD Research Institute, NYUAD, Abu Dhabi, UAE\\
\normalsize\textsuperscript{4}SDU Microelectronics, Institute of Mechanical and Electrical Engineering, University of Southern Denmark, Odense, Denmark\\
\normalsize m.terki@esi-sba.dz, \{nouhaila.innan, shao.minghao, muhammadkashif, muhammad.shafique\}@nyu.edu}, chisio@sdu.dk
\vspace{-25pt}
}

\maketitle
\thispagestyle{empty}
\pagestyle{empty}
\begin{abstract}
Large language models (LLMs) have shown strong results in classical code generation and are increasingly used to write the unit tests that safeguard that code. Whether they can do the same for quantum software is less clear, because quantum programs are probabilistic, depend on numerical tolerances, and are judged by domain-specific checkers, so a well-formed test can still disagree with the benchmark's verdict. Existing quantum benchmarks measure whether a model produces a correct solution, but not whether its tests can distinguish correct implementations from faulty ones. We introduce QTestBench, a framework that evaluates LLM-generated quantum test suites against independent benchmark checkers, never treating the tests themselves as ground truth. For each task, QTestBench prompts models to produce syntactic, semantic, and behavioral tests, injects candidate solutions at runtime, and compares each test verdict with the checker's label under three validation cohorts, each answering a distinct diagnostic question. We benchmark five LLMs on eight PennyLane QHack challenges and ten Qiskit HumanEval tasks. Generated suites almost always execute, but discrimination lags behind: the best generator reaches a test quality of 0.71 on QHack and 0.34 on Qiskit HumanEval, so executability alone overstates test quality. Oracle-verified synthetic mutants stabilize bug-detection estimates when natural faulty solutions are scarce, as on three of ten Qiskit tasks that yield none. QTestBench offers a reproducible, cross-framework benchmark for the trustworthiness of LLM-generated quantum tests.
\end{abstract}


\section{Introduction}
\label{sec:intro}
Large language models (LLMs) now turn natural-language prompts into executable code, performing strongly on classical benchmarks such as HumanEval~\cite{chen2021codex} and MBPP~\cite{austin2021mbpp}. Developers increasingly ask the same models to write the unit tests that guard this code, yet a passing generated test is not necessarily trustworthy: a suite that accepts incorrect code is worse than no test at all~\cite{wang2024llmtesting}. The risk intensifies in quantum computing, where outputs are probabilistic, tolerances matter, and correctness is often defined only through domain-specific checkers~\cite{miranskyy2019testing,garcia2023qstesting}. Benchmarks such as Qiskit HumanEval~\cite{vishwakarma2024qiskithumaneval}, QHackBench~\cite{basit2025qhackbench}, and QuanBench~\cite{guo2025quanbench} evaluate whether LLMs write correct quantum \emph{solutions}; none measures whether they write \emph{tests} that separate correct from incorrect quantum code. We close this gap with \textbf{QTestBench}.

To be useful, a generated test must (\emph{i}) execute without infrastructure failure, (\emph{ii}) accept oracle-correct candidates, and (\emph{iii}) reject oracle-wrong ones. QTestBench scores LLM-generated pytest suites on all three counts against an independent benchmark oracle, never against the tests themselves, spanning PennyLane QHack~\cite{basit2025qhackbench} and Qiskit HumanEval~\cite{vishwakarma2024qiskithumaneval}.

\subsection{Motivational Case Study}

Consider the PennyLane \emph{Coffee Conundrum} challenge and two ways an LLM-generated test can fail. The task asks for a small quantum circuit whose execution function returns a numeric answer as a string; the benchmark provides two official input/output pairs and a built-in acceptance routine. In QTestBench, an LLM first writes a pytest module for this task. We then inject \emph{candidate solutions}, a trusted human reference and LLM-generated alternatives, and ask two questions on the same input: (1)~does the \emph{generated test} pass? and (2)~does the benchmark \emph{checker} label the candidate correct? The checker is the independent ground truth; the generated test is what we evaluate. Table~\ref{tab:motivation} shows two real mismatches from our evaluation, one per test generator on this same task; in each row the test module is valid Python, yet its verdict disagrees with the checker.

\begin{table}[!t]
\caption{Two failure modes when generated tests are scored against the
benchmark checker (Coffee Conundrum challenge).}
\label{tab:motivation}
\centering
\footnotesize
\setlength{\tabcolsep}{4pt}
\renewcommand{\arraystretch}{1.15}
\begin{tabular}{@{}>{\raggedright\arraybackslash}p{0.15\columnwidth}>{\raggedright\arraybackslash}p{0.19\columnwidth}>{\raggedright\arraybackslash}p{0.21\columnwidth}cc@{}}
\toprule
\rowcolor{blue!10}
\textbf{Mode} & \textbf{Test checks} & \textbf{Probe scenario} & \textbf{Test} & \textbf{Checker} \\
\midrule
\rowcolor{orange!7}
\textbf{Too weak}\newline{\scriptsize DeepSeek V3.2~\cite{deepseek2025v32} syntactic}
& Entry-point checks only; no execution on official input
& Oracle-wrong LLM output
& \testbad{Pass} & \oracle{Wrong} \\
\rowcolor{blue!5}
\textbf{Too strict}\newline{\scriptsize GPT-5.4~\cite{openai2026gpt54} semantic}
& Numerical tolerance at $10^{-4}$
& Human reference output
& \testbad{Fail} & \oracle{Correct} \\
\bottomrule
\end{tabular}
\tablelegend{Red cells mark test verdicts that disagree with the checker; blue cells show oracle labels.}
\end{table}

In the first row, the syntactic suite never executes the candidate on official inputs, so oracle-wrong LLM solutions pass every sub-test even though their circuits return incorrect values. In the second row, the human reference passes the task acceptance routine, yet the generated semantic suite rejects it: two numerically close outputs land just outside the test's tolerance. Fig.~\ref{fig:motivation-code} lists the assertions behind each row; harness boilerplate is omitted. Neither mismatch is visible when grading only the generated \emph{code}; both surface only when grading the generated \emph{tests} against the checker, and exposing exactly such mismatches is what QTestBench is built for.

\begin{figure}[!t]
\Needspace{0.42\textheight}
\centering
\begin{codefigurebox}{Assertions behind Table~\ref{tab:motivation}}
{\footnotesize
\textit{Task.} Coffee Conundrum: candidates are loaded at runtime; official
pairs are embedded in the generated suite; the benchmark checker is the task
acceptance routine.\par\vspace{3pt}}

\begin{codesnippet}{(A) Too weak: DeepSeek V3.2 syntactic (row~1)}
{\scriptsize Structure checks only; oracle-wrong output still yields
\textbf{Test=Pass}, \textbf{Checker=Wrong}.\par}
\begin{lstlisting}[style=qpy,aboveskip=3pt,belowskip=0pt]
exec(sol, ns)
assert "run" in ns and callable(ns["run"])
assert "check" in ns and callable(ns["check"])
# no run('5') -> Pass despite wrong circuit
\end{lstlisting}
\end{codesnippet}

\begin{codesnippet}{(B) Too strict: GPT-5.4 semantic (row~2)}
{\scriptsize Human reference passes the acceptance routine; strict
\texttt{allclose} rejects numerically close outputs
(\textbf{Test=Fail}, \textbf{Checker=Correct}).\par}
\begin{lstlisting}[style=qpy,aboveskip=3pt,belowskip=0pt]
out = ns["run"]('5')          # '0.6725000000000004'
exp = '0.6735'
assert np.allclose(float(out), float(exp), atol=1e-4)
\end{lstlisting}
\end{codesnippet}
\end{codefigurebox}
\caption{Illustrative generated-test assertions for the Coffee Conundrum
challenge (Table~\ref{tab:motivation}). Each snippet shows the discriminating
checks; harness boilerplate is omitted.}
\label{fig:motivation-code}
\end{figure}

\subsection{Research Challenges}

Evaluating LLM-generated quantum tests raises four challenges: (\emph{i})~existing benchmarks grade \emph{solutions}, not \emph{tests}; (\emph{ii})~ground truth must come from an oracle independent of the generated suite; (\emph{iii})~naturally occurring wrong solutions are too sparse for stable bug-detection estimates; and (\emph{iv})~scores must transfer across SDKs with different task formats and execution models.

\subsection{Our Contributions}

We introduce \textbf{QTestBench}, a framework for measuring LLM-written
quantum test quality. Our contributions are:
\begin{itemize}
  \item \textbf{Oracle-grounded test evaluation.} QTestBench generates syntactic,
  semantic, and behavioral pytest modules per task and scores them against
  benchmark checkers that label each candidate correct, wrong, or unknown.
  \item \textbf{Discrimination metrics.} We define executability, correctness
  acceptance, bug-detection rejection, test quality, checker-input coverage, and
  composite quality for quantum test suites (Section~\ref{subsec:metrics}).
  \item \textbf{Three validation cohorts.} Configurations~A--C (Section~\ref{subsec:configs})
  isolate executability and acceptance on trusted references, discrimination
  against oracle-labeled LLM solutions, and calibrated rejection via synthetic
  mutants.
  \item \textbf{Cross-SDK evaluation.} We benchmark Claude Opus~4.6~\cite{anthropic2026claudeopus46},
  DeepSeek V3.2~\cite{deepseek2025v32}, Gemini~3 Pro~\cite{google2026gemini31pro},
  GPT-5.4~\cite{openai2026gpt54}, and Qwen3-Coder~\cite{qwen2025qwen3coder} on
  eight QHack challenges and a fixed ten-task Qiskit HumanEval scope, using the
  same metrics on PennyLane and Qiskit.
\end{itemize}

Section~\ref{sec:background} reviews related work; Section~\ref{sec:methodology}
describes the pipeline and evaluation protocol; Section~\ref{sec:results} reports
QHack and Qiskit HumanEval results; Section~\ref{sec:conclusion} concludes.

\section{Background and Related Work}
\label{sec:background}

\subsection{LLM Code Generation and Evaluation}

LLMs such as Codex~\cite{chen2021codex} and StarCoder~\cite{li2023starcoder} turn
natural-language prompts into executable code and are benchmarked on HumanEval
\cite{chen2021codex} and MBPP~\cite{austin2021mbpp}. Because judging code
automatically is itself difficult, a parallel line of work studies \emph{how}
generated code should be scored. Text-similarity metrics such as CodeBLEU
\cite{ren2020codebleu} compare program text, while execution-aware methods such
as CodeScore~\cite{dong2025codescore} predict run-time success, and FormalEval
\cite{yang2024formaleval} checks correctness without hand-written tests. Recent
reviews note that these benchmarks are fragile and often disagree
\cite{paul2024critical}. Separately, LLMs are increasingly used to author unit
tests for the code they generate~\cite{wang2024llmtesting}, but classical
benchmarks do not score whether those tests reliably discriminate correct
from incorrect implementations.

\subsection{LLM Quantum Code Generation}

Quantum programming introduces additional difficulty: outputs are probabilistic, domain-specific checkers define correctness, and tasks cover circuit construction, simulation, and hardware-oriented APIs. For IBM Qiskit, the Qiskit Code Assistant fine-tunes models for quantum programming~\cite{dupuis2024qiskitassistant}, while Qiskit HumanEval provides curated tasks with reference solutions and simulator-based checks~\cite{vishwakarma2024qiskithumaneval}. For PennyLane, QHackBench evaluates models on real QHack coding challenges drawn from hackathon-style problem sets~\cite{basit2025qhackbench} and implemented using the PennyLane framework~\cite{bergholm2018pennylane}. PennyLang introduces a PennyLane-centric dataset and studies retrieval-augmented quantum code generation~\cite{pennylang}, while PennyCoder develops lightweight domain-specific models fine-tuned for PennyLane code generation~\cite{pennycoder}. PennySynth further uses retrieval-driven data synthesis and code-aware retrieval to improve generation on QHack challenges~\cite{pennysynth}. Beyond SDK-specific studies, QuanBench covers broader quantum programming tasks and reports low accuracy for current models~\cite{guo2025quanbench}; QCoder incorporates simulator feedback~\cite{ishigaki2025qcoder}, and multi-agent pipelines iteratively repair incorrect solutions~\cite{campbell2025multiagent}. These studies share a common evaluation target: whether a model can generate correct quantum \emph{code}. However, none evaluates whether generated \emph{test suites} can reliably distinguish correct from incorrect quantum programs across multiple SDKs.

\subsection{Trusting LLM-Generated Artifacts}

Studies that inspect LLM-generated code find that high benchmark accuracy can
still hide defects, so accuracy alone is not a reliable quality
signal~\cite{sabra2024assessing,melin2024precision}. Generated artifacts should
be checked against an independent source of truth. This caution applies equally
to generated tests: a suite that passes broken code is actively misleading.
QTestBench therefore never uses generated tests as ground truth; benchmark
checkers label every candidate solution.

\subsection{Test Oracles and Mutation Testing}

Two classical software-testing ideas underpin QTestBench. A \emph{test oracle}
decides whether an output is correct; oracle selection is a long-standing
challenge~\cite{barr2015oracle}, especially for quantum code with probabilistic
outputs and numerical tolerances. \emph{Mutation testing} inserts faults into
correct programs and scores suites by mutant
detection~\cite{jia2011mutation}. QTestBench uses each benchmark's official
checker as an independent oracle and oracle-verified synthetic mutants when
natural wrong solutions are too sparse for stable bug-detection estimates.

\section{Methodology}
\label{sec:methodology}

\begin{figure*}[t]
\centering

\includegraphics[width=\textwidth,keepaspectratio]{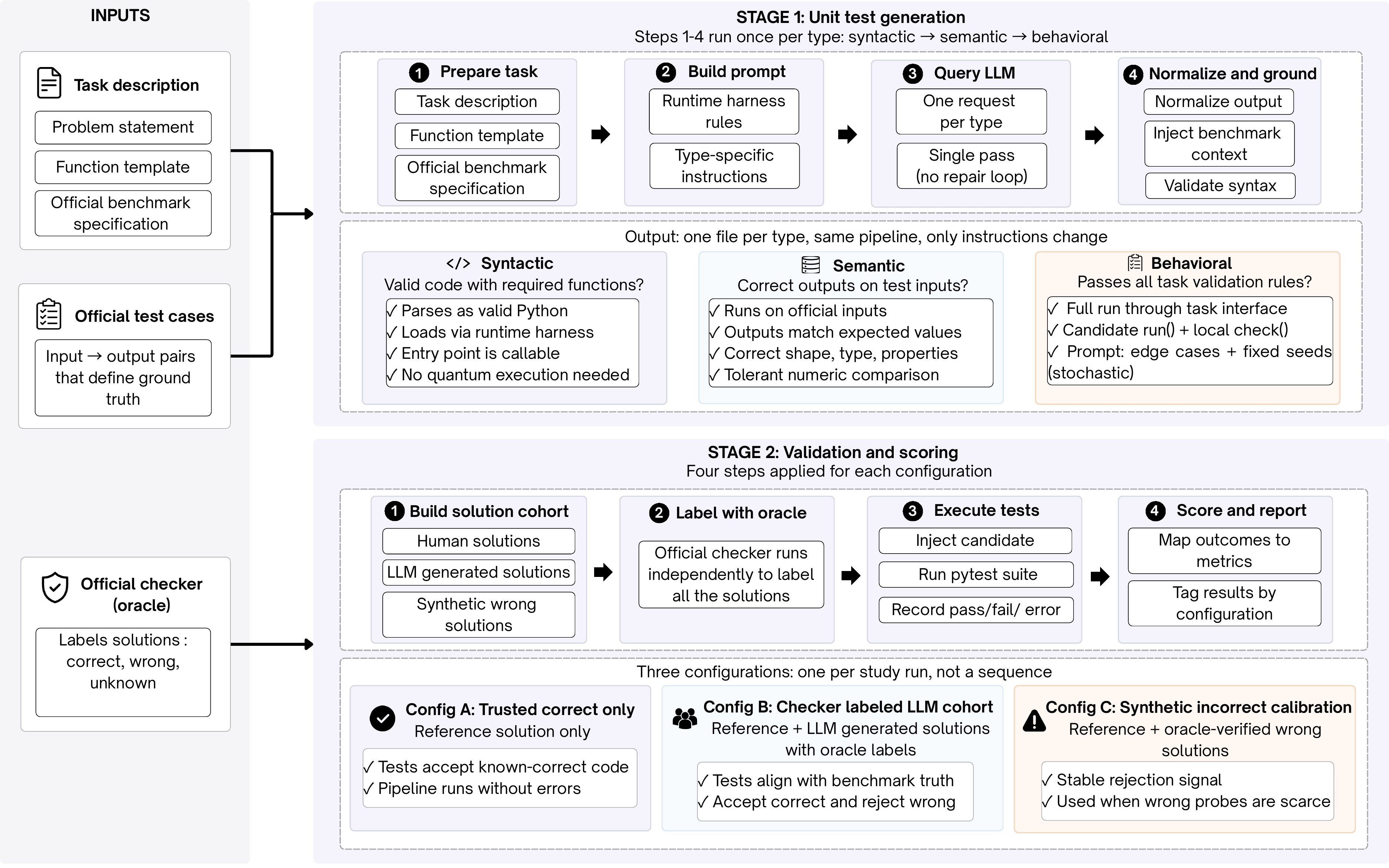}
\caption{QTestBench pipeline. \textbf{Inputs:} task description, official cases,
and benchmark checker (oracle). \textbf{Stage~1} generates syntactic, semantic,
and behavioral pytest modules (one LLM call per type, temperature $0.8$, no
repair loop). \textbf{Stage~2} builds a solution cohort, labels probes with the
oracle, executes generated tests via runtime solution injection, and reports the
metrics in Section~\ref{subsec:metrics} under Configuration~A (trusted correct only),
Configuration~B (checker-labeled LLM cohort), and Configuration~C (synthetic incorrect
calibration; Section~\ref{subsec:configs}).}
\label{fig:framework}
\end{figure*}

QTestBench evaluates LLM performance in quantum \emph{test}
generation through an automated benchmarking framework for PennyLane QHack
challenges~\cite{basit2025qhackbench} and Qiskit HumanEval tasks~\cite{vishwakarma2024qiskithumaneval}. Our methodology separates test generation
(Stage~1) from oracle-grounded validation (Stage~2): an LLM first writes pytest
modules; the framework then injects candidate solutions, runs those modules, and
compares outcomes against independent benchmark checkers. Section~\ref{subsec:metrics}
defines discrimination metrics; Section~\ref{subsec:configs} specifies three
validation cohorts; Stage~2 applies both within the pipeline in
Fig.~\ref{fig:framework} and Algorithm~\ref{alg:execution}. The same workflow
applies to both SDKs; only task materialization and oracle execution differ.

\subsection{Benchmark Tasks and Inputs}

To ensure consistency across SDKs, every evaluable task is materialized into the
same three inputs (Fig.~\ref{fig:framework}):

\begin{itemize}
  \item \textbf{Task description:} problem statement and entry-point stub that
  define the callable interface a solution must implement.
  \item \textbf{Official test cases:} published input/output pairs that ground
  semantic and behavioral assertions. Stage~1 embeds these pairs in generated
  suites; the reference solution is withheld from test-generation prompts.
  \item \textbf{Official checker (oracle):} benchmark logic that labels any
  candidate correct, wrong, or unknown. Stage~2 uses the checker only to label
  the solution cohort; generated tests never serve as ground truth.
\end{itemize}

\textbf{QHack (PennyLane).} We use eight challenges from the QHack framework
bundle, each with a template notebook, human reference submissions, and a
task-specific acceptance routine. The framework extracts prompts, stubs, and
official I/O pairs into a uniform layout shared with Stage~1 prompts and
Stage~2 oracle execution.

\textbf{Qiskit HumanEval.} Curated task records are converted into executable
benchmark units (prompt, stub, reference solution, and published checker).
Oracle execution invokes the required entry-point function and applies the
benchmark acceptance logic. All reported Qiskit results use a fixed ten-task
evaluation scope chosen before any test generation, so every model and
configuration is compared on the same tasks.

\subsection{Automated Test Generation (Stage~1)}

To ensure reproducibility, Stage~1 follows the same four-step sequence for each
test type; only the type-specific instructions in step~2 change
(Fig.~\ref{fig:framework}):
\begin{enumerate}
  \item \textbf{Prepare task.} Assemble the problem statement, entry-point stub,
  and official input/output pairs. The reference solution is excluded from the
  prompt so the model cannot copy correct answers into assertions.
  \item \textbf{Build prompt.} Combine task context with harness rules (tests
  must load injected candidate code at runtime) and instructions for syntactic,
  semantic, or behavioral checking.
  \item \textbf{Query the LLM.} Issue one request per test type at temperature
  $0.8$ with no repair or re-prompt loop.
  \item \textbf{Normalize and ground.} Strip markdown fences, validate Python
  syntax, prepend a file header, and embed official input/output pairs so
  semantic and behavioral suites compare against benchmark values.
\end{enumerate}
Three modules are saved per (task, test-generator model):
\begin{itemize}
  \item \textbf{Syntactic:} valid Python, harness load, required callable
  entry points; no circuit execution.
  \item \textbf{Semantic:} candidate execution on official cases with numerical
  comparison at $\mathrm{atol}=10^{-4}$.
  \item \textbf{Behavioral:} on QHack, call the candidate's template-local
  \texttt{run()}/\texttt{check()} helpers (harness acceptance, not the official
  checker); on Qiskit, invoke the injected entry point end-to-end.
\end{itemize}
Table~\ref{tab:test-types} and Fig.~\ref{fig:test-type-patterns} summarize how
each type interacts with the oracle on the Rainy Days forest-parity challenge.

\begin{table}[!t]
\caption{Generated test types in Stage~1: scope and oracle interaction.}
\label{tab:test-types}
\centering
\footnotesize
\setlength{\tabcolsep}{4.5pt}
\renewcommand{\arraystretch}{1.12}
\begin{tabular}{@{}>{\raggedright\arraybackslash}p{0.13\columnwidth}>{\raggedright\arraybackslash}p{0.21\columnwidth}>{\raggedright\arraybackslash}p{0.27\columnwidth}>{\raggedright\arraybackslash}p{0.23\columnwidth}@{}}
\toprule
\rowcolor{blue!10}
\textbf{Type} & \textbf{Question} & \textbf{What runs} & \textbf{Ground truth used} \\
\midrule
\rowcolor{blue!6}
\textbf{Syntactic} & Valid structure? & Parse/load candidate; no circuit execution & Required entry points only \\
\rowcolor{teal!7}
\textbf{Semantic} & Correct outputs? & Candidate execution on official cases & Embedded official I/O values \\
\rowcolor{orange!7}
\textbf{Behavioral} & Passes harness rules? & \texttt{run()} then candidate \texttt{check()} & Candidate's embedded \texttt{check()} \\
\bottomrule
\end{tabular}
\tablelegend{Row shading groups the three Stage~1 test types (see Fig.~\ref{fig:test-type-patterns}).
Behavioral suites call the candidate's template-local \texttt{check()} helper (not the
benchmark's official Stage~2 oracle).}
\end{table}

\begin{figure}[ht]
\Needspace{0.48\textheight}
\centering
\begin{codefigurebox}{Stage~1 assertion patterns (Rainy Days challenge)}
{\footnotesize
\textit{Task.} Claude Opus~4.6~\cite{anthropic2026claudeopus46} generator; parity
of forest runs on eight-qubit inputs. Only discriminating assertion bodies are
shown.\par\vspace{3pt}}

\begin{codesnippet}{Syntactic: valid structure and entry points}
\begin{lstlisting}[style=qpy,aboveskip=3pt,belowskip=0pt]
tree = ast.parse(sol)
assert "run" in funcs and "check" in funcs
exec(sol, ns)
assert callable(ns["run"]) and callable(ns["check"])
\end{lstlisting}
\end{codesnippet}

\begin{codesnippet}{Semantic: correct outputs on official inputs}
\begin{lstlisting}[style=qpy,aboveskip=3pt,belowskip=0pt]
exec(sol, ns)
for inp, exp in TEST_CASES:
    out = json.loads(ns["run"](inp))
    assert np.allclose(out, json.loads(exp), atol=1e-4)
\end{lstlisting}
\end{codesnippet}

\begin{codesnippet}{Behavioral: candidate harness \texttt{check()} (not the oracle)}
\begin{lstlisting}[style=qpy,aboveskip=3pt,belowskip=0pt]
exec(sol, ns)
for inp, exp in TEST_CASES:
    out = ns["run"](inp)
    ns["check"](out, exp)   # solution-local helper, not Stage-2 oracle
\end{lstlisting}
\end{codesnippet}
\end{codefigurebox}
\caption{Core assertion patterns for the three generated test types
(Table~\ref{tab:test-types}), exemplified on the Rainy Days challenge.}
\label{fig:test-type-patterns}
\end{figure}

\subsection{Discrimination Metrics}
\label{subsec:metrics}

Let $T$ be a generated test suite and let a \emph{validation cohort} denote the
set of candidate solutions evaluated under one configuration
(Section~\ref{subsec:configs}). For that cohort, $S_c$ is the set of
oracle-correct candidates (the trusted baseline plus any probe the oracle labels
correct) and $S_w$ is the set of oracle-wrong candidates. Candidates labeled
unknown are excluded from both pools. A suite \emph{accepts} a candidate when all
test functions pass; it \emph{rejects} when any function fails or errors.

The \textbf{executability score} (ES) is a binary indicator of whether the suite
runs without infrastructure failure on the trusted-correct baseline:
\begin{equation}
  \mathrm{ES}(T) =
  \begin{cases}
    1, & \text{if } T \text{ executes on baseline without failure,}\\
    0, & \text{otherwise.}
  \end{cases}
\end{equation}

The \textbf{correctness score} (CS) is the fraction of oracle-correct candidates
accepted by the generated tests:
\begin{equation}
  \mathrm{CS}(T) =
  \frac{\#\{s \in S_c : T \text{ accepts } s\}}{|S_c|}.
\end{equation}

The \textbf{bug-detection score} (BDS) is the fraction of oracle-wrong candidates
rejected by the generated tests, defined only when $|S_w|>0$; otherwise BDS is
\textit{n/a}:
\begin{equation}
  \mathrm{BDS}(T) =
  \frac{\#\{s \in S_w : T \text{ rejects } s\}}{|S_w|}
  \quad\text{when } |S_w|>0.
\end{equation}

The \textbf{test quality score} (TQS) combines discrimination components:
\begin{equation}
  \mathrm{TQS}(T) =
  \begin{cases}
    \mathrm{ES}\times\mathrm{CS}, & \text{Config.~A},\\
    \mathrm{ES}\times\mathrm{CS}\times\mathrm{BDS}, & |S_w|>0,\\
    \textit{n/a}, & |S_w|=0\text{ (B/C)}.
  \end{cases}
\end{equation}
Under Configuration~A, BDS is \textit{n/a} and $\mathrm{TQS}=\mathrm{ES}\times\mathrm{CS}$.
Under Configurations~B and~C, reporting BDS or TQS as zero when $|S_w|=0$ would
incorrectly penalize suites that accept all available correct probes but have no
wrong probes to reject; BDS, TQS, and CQ are therefore \textit{n/a} whenever the
cohort contains no oracle-wrong candidate. We also report \textbf{checker-input
coverage} ($C_{\mathrm{in}}$), the fraction of official checker literals present in
the generated test body, and \textbf{composite quality} (CQ),
$\mathrm{CQ}=\sqrt{\mathrm{TQS}\times C_{\mathrm{in}}}$ when TQS is defined.

\subsection{Validation Configurations}
\label{subsec:configs}

Stage~2 is repeated under three cohort compositions (Fig.~\ref{fig:framework}).
We refer to them as \textbf{Configuration~A}, \textbf{Configuration~B}, and
\textbf{Configuration~C} (Config~A--C in tables). Each configuration answers a
distinct diagnostic question while applying the metrics in
Section~\ref{subsec:metrics}:
\begin{itemize}
  \item \textbf{Configuration~A (trusted-reference cohort).} Only the
  canonical reference solution is probed. This isolates executability and
  acceptance of known-correct code; BDS is undefined and
  $\mathrm{TQS}=\mathrm{ES}\times\mathrm{CS}$.
  \item \textbf{Configuration~B (natural LLM solution cohort).} The reference
  plus oracle-labeled solutions from LLM \emph{code} generators run independently
  of test generation. This measures discrimination against naturally occurring
  wrong implementations; $\mathrm{TQS}=\mathrm{ES}\times\mathrm{CS}\times\mathrm{BDS}$
  when wrong probes exist. When $|S_w|=0$, BDS, TQS, and CQ are \textit{n/a}.
  \item \textbf{Configuration~C (synthetic mutant cohort).} The reference plus
  oracle-verified mutants of the reference, retained only when the oracle
  confirms they are wrong (\emph{synthetic incorrect calibration} in Fig.~\ref{fig:framework}).
  This supplies calibrated negatives when Configuration~B lacks wrong probes per
  task.
\end{itemize}

\subsection{Validation and Scoring (Stage~2)}

Stage~2 applies Section~\ref{subsec:configs} under the metrics in
Section~\ref{subsec:metrics} (Algorithm~\ref{alg:execution}). For each
configuration the pipeline (\emph{i})~assembles a solution cohort,
(\emph{ii})~labels non-baseline probes with the benchmark checker,
(\emph{iii})~executes generated tests with runtime candidate injection, and
(\emph{iv})~aggregates scores by test type, task, and test-generator model.

Every cohort includes a \emph{trusted baseline} $s_0$. On QHack, we select $s_0$
\emph{before} oracle-based scoring: for each challenge, every human submission
file is executed against the three Stage~1 test modules (syntactic,
semantic, behavioral); we sum the pytest sub-tests each file passes and assign
$s_0$ to the submission with the highest total (ties broken by file order). This
chooses a human reference that is compatible with the generated suites; it does
\emph{not} define ground truth. Stage~2 then applies the benchmark's
\emph{official} checker to every probe, including $s_0$, labeling each correct,
wrong, or unknown (excluded from $S_c$ and $S_w$). For each probe, the framework
injects candidate source into the pytest harness and records pass/fail outcomes.
The trusted baseline is always executed first: infrastructure failure on the
baseline (import error, timeout, or pytest crash) sets $\mathrm{ES}=0$ and zeros
ES/CS for that suite (BDS/TQS remain \textit{n/a} when $|S_w|=0$).

\begin{algorithm}[ht]
\caption{\small Unit Test Execution Harness (score aggregation follows
Section~\ref{subsec:metrics}).}
\label{alg:execution}
\small
\begin{algorithmic}[1]
\Require Generated test file $F$, candidate pool
  $\mathcal{P} = S_c \cup S_w \cup S_u$, trusted baseline $s_0 \in S_c$
\Ensure Per-candidate run records $\mathcal{R}$
\State $\mathcal{R} \leftarrow \emptyset$
\State Inject $s_0$ into harness namespace; run pytest on $F$ with timeout
\If{run fails with infrastructure error}
  \State Mark $F$ as non-executable; set $\mathrm{ES}(F) \leftarrow 0$
  \State \Return $\mathcal{R}$
\EndIf
\State Set $\mathrm{ES}(F) \leftarrow 1$
\For{each candidate $s \in \mathcal{P}$}
  \State Write $s$ to a temporary module; inject candidate source and
  entry point into the harness namespace
  \State Execute pytest on $F$ with timeout $\tau$
  \State Parse passed / failed / error counts from pytest output
  \State Classify outcome: \textsc{Accept} if all tests pass,
  \textsc{Reject} otherwise
  \State $\mathcal{R} \leftarrow \mathcal{R} \cup
  \{(F,\, s,\, \text{outcome},\, \text{trace})\}$
\EndFor
\State Aggregate $\mathcal{R}$ by test type, task, and model
\State \Return $\mathcal{R}$
\end{algorithmic}
\end{algorithm}

\subsection{Experimental Design}

\textbf{Task selection and models.} We evaluate eight QHack challenges and ten
Qiskit HumanEval tasks (0000--0003, 0019, 0020, 0024, 0034, 0035, 0063), fixed
before generation as a stratified easy/intermediate/hard mix so that every model
and configuration uses the same tasks. Five LLMs (Claude
Opus~4.6~\cite{anthropic2026claudeopus46}, DeepSeek
V3.2~\cite{deepseek2025v32}, Gemini~3 Pro~\cite{google2026gemini31pro},
GPT-5.4~\cite{openai2026gpt54}, Qwen3-Coder~\cite{qwen2025qwen3coder}) generate
tests at temperature $0.8$ (one call per type; no repair). Configuration~B uses
their separately generated solutions as probes.

\textbf{Prompts.} A fixed system prompt and type-specific user prompts request
self-contained pytest modules with at least three unit tests, runtime loading of
the candidate (never the reference solution), and
$\mathrm{atol}=10^{-4}$. Qiskit prompts also name the required entry point.
Syntactic prompts request structure checks; semantic prompts request execution on
official or specification-implied cases; behavioral prompts request end-to-end
harness checks (QHack: \texttt{run()}/\texttt{check()}; Qiskit: entry-point
call). Mentions of edge cases or fixed seeds in prompts are soft guidance for
stochastic tasks, not hard scoring requirements, and are not enforced by the
harness. 

\textbf{Mutants and sandbox.} Configuration~C applies deterministic mutations to
the trusted reference and keeps only mutants the official checker labels wrong:
\texttt{return None}; flip \texttt{==} to \texttt{!=}; remove
\texttt{measure\_all()}; replace the first CX or Hadamard with an $X$ gate;
perturb numeric literals ($1,000\rightarrow10$, $100\rightarrow1$); and, on QHack
only, force a local \texttt{check()} failure. Up to twelve candidates are tried;
the first three wrong probes are kept. Probes run in isolated temporary
directories with runtime candidate injection. On QHack, pytest limits are
180\,s per file and 60\,s per test, with 90\,s oracle labeling per candidate; on
Qiskit HumanEval the corresponding limits are 240\,s, 120\,s, and 120\,s.

\textbf{Table aggregation.} Cells average bottom-up over suite types, then
generators, then tasks (Tables~\ref{tab:qhack-per-challenge}--\ref{tab:qiskit}).
When $|S_w|=0$, BDS, TQS, and CQ are \textit{n/a} (never zero); model-level
BDS/CQ use only tasks with $|S_w|>0$. Qiskit Configuration~B omits TQS because
TQS requires BDS and three scoped tasks have $|S_w|=0$ (Case~C,
Fig.~\ref{fig:case-studies}).

Table~\ref{tab:metric-walkthrough} walks through ledger-to-score aggregation for
one Configuration~B run; Section~\ref{sec:results} then reports benchmark means.

\begin{table}[htpb]
\caption{Scoring ledger: Chalet Random Gate, Configuration~B, Claude Opus~4.6
(test generator). Probe counts then CS, BDS, and TQS (avg.\ row over test types).}
\label{tab:metric-walkthrough}
\centering
\footnotesize
\setlength{\tabcolsep}{4pt}
\renewcommand{\arraystretch}{1.08}
\begin{tabular}{@{}>{\raggedright\arraybackslash}p{0.14\columnwidth}cccccc@{}}
\toprule
\rowcolor{blue!10}
\textbf{Suite} & \textbf{ES} & \textbf{$S_c$} & \textbf{CS} &
\textbf{$S_w$} & \textbf{BDS} & \textbf{TQS} \\
\midrule
\rowcolor{blue!6}
\textbf{Syntactic} & 1 & 6/6 & 1.00 & 2/3 & 0.67 &
\cellcolor{yellow!22}0.67 \\
\rowcolor{teal!7}
\textbf{Semantic} & 1 & 6/6 & 1.00 & 3/3 & 1.00 &
\cellcolor{green!22}1.00 \\
\rowcolor{orange!7}
\textbf{Behavioral} & 1 & 6/6 & 1.00 & 3/3 & 1.00 &
\cellcolor{green!22}1.00 \\
\midrule
\textbf{Avg.} & -- & -- & 1.00 & -- & 0.89 &
\cellcolor{green!22}\textbf{0.89} \\
\bottomrule
\end{tabular}
\tablelegend{$S_c$/$S_w$: oracle-labeled probe pass/tot and fail/tot counts;
TQS $=\mathrm{ES}\times\mathrm{CS}\times\mathrm{BDS}$. Row colors match
Table~\ref{tab:test-types}; TQS shading matches result tables.}
\end{table}

\section{Results and Discussion}
\label{sec:results}

This section asks whether each suite runs and accepts correct code~(A), catches
bugs models actually make~(B), and catches injected faults~(C). Metrics follow
Sections~\ref{subsec:metrics} and~\ref{subsec:configs}. Across both benchmarks,
models write tests that run, but are better at accepting correct code than at
rejecting incorrect code.

\subsection{QHack Results}

Table~\ref{tab:qhack-per-challenge} reports per-challenge means averaged over the
five test generators; Table~\ref{tab:qhack} aggregates the same run by model.

\emph{Tests run reliably, but acceptance already separates models.}
Under Configuration~A, executability is at least 0.79 for every model and reaches
1.00 for Claude Opus~4.6, Gemini~3
Pro, and GPT-5.4, so runtime
injection is reliable. Acceptance already spreads the field: Gemini~3 Pro
accepts almost all correct code ($\mathrm{CS}=\mathrm{TQS}=0.96$), ahead of
Claude Opus~4.6 (0.88) and GPT-5.4 (0.83), whereas DeepSeek
V3.2 (0.42) and Qwen3-Coder
(0.37) write tests so strict that they reject much of the trusted reference
before any bug is considered. The Coffee Conundrum row ($\mathrm{CS}=0.47$;
Table~\ref{tab:motivation}, Fig.~\ref{fig:motivation-code}) shows that high
executability alone is insufficient when semantic checks disagree with the
benchmark checker.

\emph{Bug detection, not acceptance, decides overall quality.}
Configurations~B and~C add oracle-wrong probes, so BDS contributes to TQS and
reorders the ranking. Claude Opus~4.6 attains the highest test quality
($\mathrm{TQS}=0.69$ and $0.71$) by pairing strong acceptance with the best bug
detection (BDS $=0.80$ and $0.83$). Gemini~3 Pro keeps the highest acceptance
($\mathrm{CS}=0.96$) but detects fewer bugs (BDS $=0.67$), so its
Configuration~A lead does not survive once rejection matters. Configurations~B
and~C agree closely for every model, indicating that synthetic mutants reproduce
the discrimination signal of naturally occurring wrong solutions on QHack.

\emph{High coverage does not imply good tests.}
Coffee embeds almost all checker inputs ($C_{\mathrm{in}}=0.97$) yet reaches
only $\mathrm{TQS}=0.13$, so coverage without discrimination is not enough.

\begin{table*}[!t]
\caption{QHack benchmark: mean per-challenge scores (averaged over five test
generators). Configs~A--C appear as column groups; see legend below for
TQS/CQ shading.}
\label{tab:qhack-per-challenge}
\centering
\footnotesize
\renewcommand{\arraystretch}{1.0}
\setlength{\tabcolsep}{3.2pt}
\begin{tabular}{@{} L{0.108\textwidth} *{5}{c} @{\hspace{4pt}} *{6}{c} @{\hspace{4pt}} *{6}{c} @{}}
\toprule
& \multicolumn{5}{c}{\textbf{Config~A}} & \multicolumn{6}{c}{\textbf{Config~B}} & \multicolumn{6}{c}{\textbf{Config~C}} \\
\cmidrule(lr){2-6}\cmidrule(lr){7-12}\cmidrule(lr){13-18}
\textbf{Challenge} & ES & CS & TQS & $C_{\mathrm{in}}$ & CQ & ES & CS & BDS & TQS & $C_{\mathrm{in}}$ & CQ & ES & CS & BDS & TQS & $C_{\mathrm{in}}$ & CQ \\
\midrule
\scriptsize QSP swamp & 0.93 & 0.73 & \cellcolor{yellow!22}0.73 & 0.50 & \cellcolor{orange!16}0.47 & 0.93 & 0.73 & 0.73 & \cellcolor{yellow!22}0.53 & 0.50 & \cellcolor{orange!16}0.47 & 0.93 & 0.73 & 0.73 & \cellcolor{yellow!22}0.53 & 0.50 & \cellcolor{orange!16}0.47 \\
\scriptsize Chalet gate & 0.87 & 0.80 & \cellcolor{green!22}\textbf{0.80} & 0.60 & \cellcolor{yellow!22}\textbf{0.53} & 0.87 & 0.80 & 0.64 & \cellcolor{yellow!22}0.58 & 0.60 & \cellcolor{yellow!22}0.53 & 0.87 & 0.80 & 0.67 & \cellcolor{yellow!22}0.60 & 0.60 & \cellcolor{yellow!22}0.53 \\
\scriptsize Coffee & 1.00 & 0.47 & \cellcolor{orange!16}0.47 & 0.97 & \cellcolor{orange!16}0.45 & 1.00 & 0.47 & 0.67 & \cellcolor{red!14}0.13 & 0.97 & \cellcolor{red!14}0.13 & 1.00 & 0.47 & 0.67 & \cellcolor{red!14}0.13 & 0.97 & \cellcolor{red!14}0.13 \\
\scriptsize Contextuality & 0.93 & 0.80 & \cellcolor{green!22}\textbf{0.80} & 0.47 & \cellcolor{orange!16}\textbf{0.40} & 0.93 & 0.80 & 0.60 & \cellcolor{orange!16}0.47 & 0.47 & \cellcolor{orange!16}0.40 & 0.93 & 0.80 & 0.60 & \cellcolor{orange!16}0.47 & 0.47 & \cellcolor{orange!16}0.40 \\
\scriptsize Mach--Zehnder & 0.93 & 0.80 & \cellcolor{green!22}\textbf{0.80} & 0.53 & \cellcolor{orange!16}\textbf{0.47} & 0.93 & 0.80 & 0.60 & \cellcolor{orange!16}0.47 & 0.53 & \cellcolor{orange!16}0.47 & 0.93 & 0.80 & 0.60 & \cellcolor{orange!16}0.47 & 0.53 & \cellcolor{orange!16}0.47 \\
\scriptsize Rainy days & 1.00 & 0.40 & \cellcolor{orange!16}0.40 & 0.50 & \cellcolor{orange!16}0.31 & 1.00 & 0.44 & 0.93 & \cellcolor{orange!16}0.37 & 0.50 & \cellcolor{orange!16}0.40 & 1.00 & 0.40 & 0.93 & \cellcolor{orange!16}0.33 & 0.50 & \cellcolor{orange!16}0.31 \\
\scriptsize Eigentracks & 1.00 & 0.87 & \cellcolor{green!22}\textbf{0.87} & 0.53 & \cellcolor{yellow!22}\textbf{0.53} & 1.00 & 0.87 & 0.71 & \cellcolor{yellow!22}0.57 & 0.53 & \cellcolor{yellow!22}0.53 & 1.00 & 0.87 & 0.73 & \cellcolor{yellow!22}0.60 & 0.53 & \cellcolor{yellow!22}0.53 \\
\scriptsize Triple-H & 0.93 & 0.67 & \cellcolor{yellow!22}0.67 & 0.53 & \cellcolor{orange!16}0.40 & 0.93 & 0.73 & 0.60 & \cellcolor{orange!16}0.40 & 0.53 & \cellcolor{orange!16}0.40 & 0.93 & 0.67 & 0.60 & \cellcolor{orange!16}0.40 & 0.53 & \cellcolor{orange!16}0.40 \\
\bottomrule
\end{tabular}
\par\vspace{5pt}\begin{minipage}{\textwidth}\scriptsize\textit{Legend:} TQS and CQ cell shading (higher is better): \colorbox{green!22}{\rule{0pt}{0.75em}\hspace{0.9em}}${\geq}0.75$; \colorbox{yellow!22}{\rule{0pt}{0.75em}\hspace{0.9em}}${\geq}0.50$; \colorbox{orange!16}{\rule{0pt}{0.75em}\hspace{0.9em}}${\geq}0.25$; \colorbox{red!14}{\rule{0pt}{0.75em}\hspace{0.9em}}${<}0.25$; \colorbox{gray!8}{\rule{0pt}{0.75em}\hspace{0.9em}}\textit{n/a} = BDS/TQS/CQ undefined ($|S_w|=0$).\end{minipage}
\end{table*}

Table~\ref{tab:qhack} aggregates the same run by test-generator model.

\begin{table*}[!t]
\caption{QHack benchmark: mean scores per test-generator model (Configs~A--C
as column groups; best TQS per configuration in bold). Layout matches
Table~\ref{tab:qhack-per-challenge}.}
\label{tab:qhack}
\centering
\footnotesize
\renewcommand{\arraystretch}{1.0}
\setlength{\tabcolsep}{3.5pt}
\begin{tabular}{@{} L{0.12\textwidth} *{5}{c} @{\hspace{4pt}} *{6}{c} @{\hspace{4pt}} *{6}{c} @{}}
\toprule
& \multicolumn{5}{c}{\textbf{Config~A}} & \multicolumn{6}{c}{\textbf{Config~B}} & \multicolumn{6}{c}{\textbf{Config~C}} \\
\cmidrule(lr){2-6}\cmidrule(lr){7-12}\cmidrule(lr){13-18}
\textbf{Model} & ES & CS & TQS & $C_{\mathrm{in}}$ & CQ & ES & CS & BDS & TQS & $C_{\mathrm{in}}$ & CQ & ES & CS & BDS & TQS & $C_{\mathrm{in}}$ & CQ \\
\midrule
\scriptsize Claude Opus 4.6 & 1.00 & 0.88 & \cellcolor{green!22}0.88 & 0.69 & \cellcolor{yellow!22}0.57 & 1.00 & 0.88 & 0.80 & \cellcolor{yellow!22}\textbf{0.69} & 0.69 & \cellcolor{yellow!22}\textbf{0.55} & 1.00 & 0.88 & 0.83 & \cellcolor{yellow!22}\textbf{0.71} & 0.69 & \cellcolor{yellow!22}\textbf{0.53} \\
\scriptsize DeepSeek V3.2 & 0.96 & 0.42 & \cellcolor{orange!16}0.42 & 0.35 & \cellcolor{orange!16}0.28 & 0.96 & 0.47 & 0.75 & \cellcolor{orange!16}0.26 & 0.35 & \cellcolor{orange!16}0.27 & 0.96 & 0.42 & 0.75 & \cellcolor{red!14}0.25 & 0.35 & \cellcolor{red!14}0.25 \\
\scriptsize Gemini 3 Pro & 1.00 & 0.96 & \cellcolor{green!22}\textbf{0.96} & 0.71 & \cellcolor{yellow!22}\textbf{0.67} & 1.00 & 0.96 & 0.67 & \cellcolor{yellow!22}0.63 & 0.71 & \cellcolor{yellow!22}0.63 & 1.00 & 0.96 & 0.67 & \cellcolor{yellow!22}0.63 & 0.71 & \cellcolor{yellow!22}0.63 \\
\scriptsize GPT-5.4 & 1.00 & 0.83 & \cellcolor{green!22}0.83 & 0.71 & \cellcolor{yellow!22}0.58 & 1.00 & 0.84 & 0.71 & \cellcolor{yellow!22}0.55 & 0.71 & \cellcolor{yellow!22}0.56 & 1.00 & 0.83 & 0.71 & \cellcolor{yellow!22}0.54 & 0.71 & \cellcolor{yellow!22}0.54 \\
\scriptsize Qwen3-Coder & 0.79 & 0.37 & \cellcolor{orange!16}0.37 & 0.44 & \cellcolor{red!14}0.12 & 0.79 & 0.37 & 0.50 & \cellcolor{red!14}0.08 & 0.44 & \cellcolor{red!14}0.08 & 0.79 & 0.37 & 0.50 & \cellcolor{red!14}0.08 & 0.44 & \cellcolor{red!14}0.08 \\
\bottomrule
\end{tabular}
\par\vspace{5pt}\begin{minipage}{\textwidth}\scriptsize\textit{Legend:} TQS and CQ cell shading (higher is better): \colorbox{green!22}{\rule{0pt}{0.75em}\hspace{0.9em}}${\geq}0.75$; \colorbox{yellow!22}{\rule{0pt}{0.75em}\hspace{0.9em}}${\geq}0.50$; \colorbox{orange!16}{\rule{0pt}{0.75em}\hspace{0.9em}}${\geq}0.25$; \colorbox{red!14}{\rule{0pt}{0.75em}\hspace{0.9em}}${<}0.25$; \colorbox{gray!8}{\rule{0pt}{0.75em}\hspace{0.9em}}\textit{n/a} = BDS/TQS/CQ undefined ($|S_w|=0$).\end{minipage}
\end{table*}

\subsection{Qiskit HumanEval Results}

Tables~\ref{tab:qiskit-per-task}--\ref{tab:qiskit} report the fixed ten-task
scope, organized around the same three questions.

\emph{The protocol transfers, but tasks vary widely.}
Under Configuration~A, harness injection carries over unchanged (four models
reach $\mathrm{ES}=1.00$; Qwen3-Coder averages 0.93), and Claude Opus~4.6 again
leads on acceptance ($\mathrm{CS}=\mathrm{TQS}=0.77$). What changes is the
spread across tasks: the Bell-state simulator reaches $\mathrm{TQS}=0.87$
($C_{\mathrm{in}}=0.93$) while GHZ-state construction collapses to
$\mathrm{TQS}=0.13$ at comparable coverage ($C_{\mathrm{in}}=0.78$), because
semantic suites sometimes echo checker literals without accepting the reference.

\emph{Realistic bugs are caught only moderately, and sometimes cannot be
measured.}
Configuration~B enriches the pool with LLM solutions (199 probes per run:
119 correct, 80 wrong). Bug detection is middling (BDS $=0.45$--$0.60$):
Qwen3-Coder rejects the most wrong code (0.60) while Claude Opus~4.6 accepts the
most correct code ($\mathrm{CS}=0.84$), again the acceptance--rejection
trade-off. Three of the ten tasks attract no wrong solutions and report
\textit{n/a} for BDS, TQS, and CQ, so the cohort, not the suite, is limiting
(Case~C). Because TQS is defined as $\mathrm{ES}\times\mathrm{CS}\times\mathrm{BDS}$
only when $|S_w|>0$, Table~\ref{tab:qiskit-per-task} omits the TQS column for
Configuration~B rather than reporting zeros.

\emph{Synthetic faults restore measurability but expose low ceilings.}
Configuration~C injects verified mutants into every task, so BDS is always
defined: Gemini~3 Pro then leads ($\mathrm{TQS}=0.34$), with Claude Opus~4.6 at
0.29. These ceilings sit well below QHack because many Qiskit checkers contain
few extractable literals (Deutsch--Jozsa has $C_{\mathrm{in}}=0.00$), pushing CQ
toward zero even when acceptance is moderate.

\begin{table*}[!t]
\caption{Qiskit HumanEval: mean per-task scores (ten-task scope; averaged over
five test generators). Configuration~A: ES, CS, TQS
($\mathrm{TQS}=\mathrm{ES}\times\mathrm{CS}$), $C_{\mathrm{in}}$, CQ.
Configuration~B: ES, CS, BDS (\textit{n/a} if no oracle-wrong probes),
$C_{\mathrm{in}}$, CQ. Configuration~C: full metric set.
Rows list short HumanEval task identifiers.}
\label{tab:qiskit-per-task}
\centering
\footnotesize
\renewcommand{\arraystretch}{1.0}
\setlength{\tabcolsep}{3.5pt}
\begin{tabular}{@{} L{0.108\textwidth} *{5}{c} @{\hspace{4pt}} *{5}{c} @{\hspace{4pt}} *{6}{c} @{}}
\toprule
& \multicolumn{5}{c}{\textbf{Config~A}} & \multicolumn{5}{c}{\textbf{Config~B}} & \multicolumn{6}{c}{\textbf{Config~C}} \\
\cmidrule(lr){2-6}\cmidrule(lr){7-11}\cmidrule(lr){12-17}
\textbf{Task} & ES & CS & TQS & $C_{\mathrm{in}}$ & CQ & ES & CS & BDS & $C_{\mathrm{in}}$ & CQ & ES & CS & BDS & TQS & $C_{\mathrm{in}}$ & CQ \\
\midrule
\parbox[t]{0.108\textwidth}{\scriptsize\raggedright \texttt{create\_quantum}\\[-1pt]\texttt{circuit}} & 1.00 & 0.73 & \cellcolor{yellow!22}0.73 & 0.00 & \cellcolor{red!14}0.00 & 1.00 & 0.70 & \cellcolor{gray!8}{\scriptsize\textit{n/a}} & 0.00 & \cellcolor{gray!8}{\scriptsize\textit{n/a}} & 1.00 & 0.73 & \cellcolor{green!22}1.00 & \cellcolor{yellow!22}0.73 & 0.00 & \cellcolor{red!14}0.00 \\
\parbox[t]{0.108\textwidth}{\scriptsize\raggedright \texttt{run\_bell\_state}\\[-1pt]\texttt{simulator}} & 1.00 & 0.87 & \cellcolor{green!22}\textbf{0.87} & 0.93 & \cellcolor{green!22}\textbf{0.83} & 1.00 & 0.87 & \cellcolor{yellow!22}0.73 & 0.93 & \cellcolor{yellow!22}0.60 & 1.00 & 0.93 & \cellcolor{yellow!22}0.62 & \cellcolor{yellow!22}0.56 & 0.93 & \cellcolor{yellow!22}0.59 \\
\parbox[t]{0.108\textwidth}{\scriptsize\raggedright \texttt{create\_bell}\\[-1pt]\texttt{statevector}} & 1.00 & 0.73 & \cellcolor{yellow!22}0.73 & 0.93 & \cellcolor{yellow!22}0.71 & 1.00 & 0.73 & \cellcolor{yellow!22}0.73 & 0.93 & \cellcolor{orange!16}0.47 & 1.00 & 0.73 & \cellcolor{red!14}0.00 & \cellcolor{red!14}0.00 & 0.93 & \cellcolor{red!14}0.00 \\
\parbox[t]{0.108\textwidth}{\scriptsize\raggedright \texttt{create\_ghz}} & 0.93 & 0.13 & \cellcolor{red!14}0.13 & 0.78 & \cellcolor{red!14}0.11 & 0.93 & 0.80 & \cellcolor{gray!8}{\scriptsize\textit{n/a}} & 0.78 & \cellcolor{gray!8}{\scriptsize\textit{n/a}} & 0.93 & 0.13 & \cellcolor{green!22}0.83 & \cellcolor{red!14}0.03 & 0.78 & \cellcolor{red!14}0.04 \\
\parbox[t]{0.108\textwidth}{\scriptsize\raggedright \texttt{transpile}\\[-1pt]\texttt{circuit\_maxopt}} & 1.00 & 0.40 & \cellcolor{orange!16}0.40 & 0.00 & \cellcolor{red!14}0.00 & 1.00 & 0.45 & \cellcolor{gray!8}{\scriptsize\textit{n/a}} & 0.00 & \cellcolor{gray!8}{\scriptsize\textit{n/a}} & 1.00 & 0.40 & \cellcolor{red!14}0.00 & \cellcolor{red!14}0.00 & 0.00 & \cellcolor{red!14}0.00 \\
\parbox[t]{0.108\textwidth}{\scriptsize\raggedright \texttt{transpile\_ghz}\\[-1pt]\texttt{customlayout}} & 1.00 & 0.27 & \cellcolor{orange!16}0.27 & 0.00 & \cellcolor{red!14}0.00 & 1.00 & 0.27 & \cellcolor{green!22}0.87 & 0.00 & \cellcolor{red!14}0.00 & 1.00 & 0.27 & \cellcolor{red!14}0.00 & \cellcolor{red!14}0.00 & 0.00 & \cellcolor{red!14}0.00 \\
\parbox[t]{0.108\textwidth}{\scriptsize\raggedright \texttt{dj\_algorithm}} & 1.00 & 0.47 & \cellcolor{orange!16}0.47 & 0.00 & \cellcolor{red!14}0.00 & 1.00 & 0.79 & \cellcolor{green!22}0.88 & 0.00 & \cellcolor{red!14}0.00 & 1.00 & 0.47 & \cellcolor{green!22}0.77 & \cellcolor{red!14}0.23 & 0.00 & \cellcolor{red!14}0.00 \\
\parbox[t]{0.108\textwidth}{\scriptsize\raggedright \texttt{run\_jobs\_on}\\[-1pt]\texttt{batch}} & 1.00 & 0.47 & \cellcolor{orange!16}0.47 & 0.65 & \cellcolor{orange!16}0.30 & 1.00 & 0.47 & \cellcolor{yellow!22}0.60 & 0.65 & \cellcolor{red!14}0.13 & 1.00 & 0.47 & \cellcolor{yellow!22}0.53 & \cellcolor{red!14}0.00 & 0.65 & \cellcolor{red!14}0.00 \\
\parbox[t]{0.108\textwidth}{\scriptsize\raggedright \texttt{run\_circuit}\\[-1pt]\texttt{with\_dd\_trex}} & 0.93 & 0.47 & \cellcolor{orange!16}0.47 & 0.00 & \cellcolor{red!14}0.00 & 0.93 & 0.47 & \cellcolor{yellow!22}0.66 & 0.00 & \cellcolor{red!14}0.00 & 0.93 & 0.47 & \cellcolor{yellow!22}0.67 & \cellcolor{red!14}0.20 & 0.00 & \cellcolor{red!14}0.00 \\
\parbox[t]{0.108\textwidth}{\scriptsize\raggedright \texttt{bb84\_circuit}\\[-1pt]\texttt{generate\_key}} & 1.00 & 0.53 & \cellcolor{yellow!22}0.53 & 0.00 & \cellcolor{red!14}0.00 & 1.00 & 0.51 & \cellcolor{yellow!22}0.70 & 0.00 & \cellcolor{red!14}0.00 & 1.00 & 0.53 & \cellcolor{yellow!22}0.70 & \cellcolor{red!14}0.23 & 0.00 & \cellcolor{red!14}0.00 \\
\bottomrule
\end{tabular}
\par\vspace{5pt}\begin{minipage}{\textwidth}\scriptsize\textit{Legend:} TQS and CQ cell shading (higher is better): \colorbox{green!22}{\rule{0pt}{0.75em}\hspace{0.9em}}${\geq}0.75$; \colorbox{yellow!22}{\rule{0pt}{0.75em}\hspace{0.9em}}${\geq}0.50$; \colorbox{orange!16}{\rule{0pt}{0.75em}\hspace{0.9em}}${\geq}0.25$; \colorbox{red!14}{\rule{0pt}{0.75em}\hspace{0.9em}}${<}0.25$; \colorbox{gray!8}{\rule{0pt}{0.75em}\hspace{0.9em}}\textit{n/a} = metric undefined ($|S_w|=0$); Config~B has no TQS column because TQS requires BDS and three tasks have $|S_w|=0$.\end{minipage}
\end{table*}

\begin{table*}[!t]
\caption{Qiskit HumanEval: mean scores per test-generator model over the scoped
ten-task runs. Same column layout as Table~\ref{tab:qiskit-per-task}; best TQS per
configuration in bold (Config~A and Config~C).}
\label{tab:qiskit}
\centering
\footnotesize
\renewcommand{\arraystretch}{1.0}
\setlength{\tabcolsep}{3.2pt}
\begin{tabular}{@{} L{0.12\textwidth} *{5}{c} @{\hspace{4pt}} *{5}{c} @{\hspace{4pt}} *{6}{c} @{}}
\toprule
& \multicolumn{5}{c}{\textbf{Config~A}} & \multicolumn{5}{c}{\textbf{Config~B}} & \multicolumn{6}{c}{\textbf{Config~C}} \\
\cmidrule(lr){2-6}\cmidrule(lr){7-11}\cmidrule(lr){12-17}
\textbf{Model} & ES & CS & TQS & $C_{\mathrm{in}}$ & CQ & ES & CS & BDS & $C_{\mathrm{in}}$ & CQ & ES & CS & BDS & TQS & $C_{\mathrm{in}}$ & CQ \\
\midrule
\scriptsize Claude Opus 4.6 & 1.00 & 0.77 & \cellcolor{green!22}\textbf{0.77} & 0.36 & \cellcolor{orange!16}\textbf{0.28} & 1.00 & 0.84 & \cellcolor{yellow!22}0.65 & 0.36 & \cellcolor{orange!16}0.27 & 1.00 & 0.77 & \cellcolor{orange!16}0.46 & \cellcolor{orange!16}0.29 & 0.36 & \cellcolor{red!14}0.07 \\
\scriptsize DeepSeek V3.2 & 1.00 & 0.30 & \cellcolor{orange!16}0.30 & 0.37 & \cellcolor{red!14}0.13 & 1.00 & 0.42 & \cellcolor{yellow!22}0.73 & 0.37 & \cellcolor{red!14}0.07 & 1.00 & 0.33 & \cellcolor{yellow!22}0.52 & \cellcolor{red!14}0.06 & 0.37 & \cellcolor{red!14}0.03 \\
\scriptsize Gemini 3 Pro & 1.00 & 0.70 & \cellcolor{yellow!22}0.70 & 0.33 & \cellcolor{red!14}0.24 & 1.00 & 0.82 & \cellcolor{yellow!22}0.75 & 0.33 & \cellcolor{red!14}0.19 & 1.00 & 0.70 & \cellcolor{yellow!22}0.54 & \cellcolor{orange!16}\textbf{0.34} & 0.33 & \cellcolor{red!14}\textbf{0.09} \\
\scriptsize GPT-5.4 & 1.00 & 0.47 & \cellcolor{orange!16}0.47 & 0.33 & \cellcolor{red!14}0.16 & 1.00 & 0.57 & \cellcolor{yellow!22}0.72 & 0.33 & \cellcolor{red!14}0.14 & 1.00 & 0.47 & \cellcolor{orange!16}0.48 & \cellcolor{red!14}0.14 & 0.33 & \cellcolor{red!14}0.05 \\
\scriptsize Qwen3-Coder & 0.93 & 0.30 & \cellcolor{orange!16}0.30 & 0.26 & \cellcolor{red!14}0.16 & 0.93 & 0.37 & \cellcolor{green!22}0.85 & 0.26 & \cellcolor{red!14}0.18 & 0.93 & 0.30 & \cellcolor{yellow!22}0.56 & \cellcolor{red!14}0.16 & 0.26 & \cellcolor{red!14}0.08 \\
\bottomrule
\end{tabular}
\par\vspace{5pt}\begin{minipage}{\textwidth}\scriptsize\textit{Legend:} TQS and CQ cell shading (higher is better): \colorbox{green!22}{\rule{0pt}{0.75em}\hspace{0.9em}}${\geq}0.75$; \colorbox{yellow!22}{\rule{0pt}{0.75em}\hspace{0.9em}}${\geq}0.50$; \colorbox{orange!16}{\rule{0pt}{0.75em}\hspace{0.9em}}${\geq}0.25$; \colorbox{red!14}{\rule{0pt}{0.75em}\hspace{0.9em}}${<}0.25$; \colorbox{gray!8}{\rule{0pt}{0.75em}\hspace{0.9em}}\textit{n/a} = metric undefined ($|S_w|=0$); Config~B has no TQS column because TQS requires BDS and three tasks have $|S_w|=0$.\end{minipage}
\end{table*}
\begin{figure}[htpb]
\centering
\begin{codefigurebox}{Illustrative case studies: metrics in practice}
{\footnotesize
Each panel reports probe verdicts (\emph{Test} vs.\ benchmark \emph{Checker})
and the resulting ES/CS/BDS/TQS scores under the stated configuration.\par\vspace{3pt}}

\begin{casestudy}{Case A: Too weak (missed bugs; Configuration~B)}
{\scriptsize
\textbf{Challenge} Mach--Zehnder Cabin \textbar\
\textbf{Generator} GPT-5.4 syntactic\par\vspace{2pt}}
\setlength{\tabcolsep}{4pt}
\scriptsize
\begin{tabular}{@{}lcc@{}}
\toprule
\textbf{Candidate probe} & \textbf{Test} & \textbf{Checker} \\
\midrule
Oracle-wrong LLM solution & Pass & Wrong \\
Human baseline & Pass & Correct \\
\bottomrule
\end{tabular}
\casemetarow{1}{1}{0}{0}
\casetakeaway{$\mathrm{ES}\times\mathrm{CS}=1$ masks zero bug detection; BDS is required.}
\end{casestudy}

\begin{casestudy}{Case B: Too strict (false reject; Configuration~B)}
{\scriptsize
\textbf{Challenge} Contextuality Dunes \textbar\
\textbf{Generator} DeepSeek V3.2 semantic\par\vspace{2pt}}
\setlength{\tabcolsep}{4pt}
\scriptsize
\begin{tabular}{@{}lcc@{}}
\toprule
\textbf{Candidate probe} & \textbf{Test} & \textbf{Checker} \\
\midrule
Human reference & Fail & Correct \\
Oracle-wrong LLM solutions & Fail & Wrong \\
\bottomrule
\end{tabular}
\casemetarow{1}{0}{1}{0}
\casetakeaway{High BDS alone is insufficient when CS rejects trusted code.}
\end{casestudy}

\begin{casestudy}{Case C: Cohort design (GHZ construction; Config.\ B vs.\ C)}
{\scriptsize
\textbf{Generator} Claude Opus~4.6 \textbar\
same suites, different probe pools (Table~\ref{tab:qiskit-per-task}).\par\vspace{2pt}}
\setlength{\tabcolsep}{3pt}
\scriptsize
\begin{tabular}{@{}lcccccc@{}}
\toprule
\textbf{Config.} & \textbf{Wrong probes} & \textbf{ES} & \textbf{CS} & \textbf{BDS} & \textbf{TQS} \\
\midrule
B (LLM cohort) & 0 ($|S_w|=0$) & 0.93 & 0.80 & n/a & n/a \\
C (synthetic) & oracle-wrong mutants & 0.93 & 0.13 & 0.83 & 0.03 \\
\bottomrule
\end{tabular}
\casetakeaway{With $|S_w|=0$, BDS/TQS are \textit{n/a}; Configuration~C restores
measurable rejection, but low CS caps TQS at 0.03.}
\end{casestudy}
\end{codefigurebox}
\caption{Three case studies on challenges other than the motivational example,
connecting QTestBench metrics to concrete oracle vs.\ test outcomes.}
\label{fig:case-studies}
\end{figure}

\subsection{Metric Case Studies}
\label{subsec:case-studies}

The aggregate tables show that discrimination is the bottleneck;
Fig.~\ref{fig:case-studies} illustrates why on challenges other than
Section~\ref{sec:intro}.

\emph{Case~A: a perfect-looking suite can be worthless.}
On Mach--Zehnder Cabin, a GPT-5.4~\cite{openai2026gpt54} syntactic suite accepts
every correct probe (ES $=$ CS $=1$) but also passes oracle-wrong solutions
never run on official inputs (BDS $=0$), so TQS collapses to zero.

\emph{Case~B: the opposite failure is just as damaging.}
On Contextuality Dunes, a DeepSeek V3.2~\cite{deepseek2025v32} suite rejects
even the correct human reference (BDS $=1$, CS $=0$, TQS $=0$). Cases~A and~B
are mirror images, so acceptance and rejection must be reported separately.

\emph{Case~C: sometimes the cohort, not the suite, sets the score.}
On \texttt{create\_ghz}, the same Claude Opus~4.6 suites give BDS/TQS
\textit{n/a} under Configuration~B ($|S_w|=0$) but BDS $=0.83$ under
Configuration~C; low CS ($0.13$) caps TQS at 0.03, which is why Qiskit
Configuration~B omits a pooled TQS column.

\subsection{Cross-Framework Synthesis}

The same metric definitions transfer across PennyLane and Qiskit without changing
the protocol. Executability rarely limits ($\mathrm{ES}\geq0.93$ for most
pairs), while CS and BDS separate generators. Claude Opus~4.6 and Gemini~3 Pro
lead on QHack ($\mathrm{TQS}=0.71$ and $0.63$ under Configuration~C); Qiskit
peaks lower (0.34, Gemini), reflecting sparser wrong probes and fewer
extractable checker literals. Absolute TQS depends on probe composition, so BDS
should be read with probe counts and as \textit{n/a} when $|S_w|=0$.

On QHack, Configurations~B and~C agree closely, so synthetic mutants reproduce
the natural cohort signal. On Qiskit, Configuration~B leaves three tasks with
$|S_w|=0$ (no pooled TQS), and Configuration~C restores defined BDS with only
moderate TQS. High $C_{\mathrm{in}}$ alone is insufficient: Coffee reaches
$C_{\mathrm{in}}=0.97$ with $\mathrm{TQS}=0.13$, the same too-weak pattern
illustrated by Case~A on Mach--Zehnder Cabin (high acceptance without bug
detection). Even under Configuration~C, single-pass generation remains limited;
refinement with execution feedback~\cite{basit2025qhackbench} is a natural next
step.
\emph{Limitations.} The evaluation is limited to a fixed
temperature of $0.8$, a single generation pass without
repair, ten selected Qiskit tasks, and benchmark-specific
checker oracles.

\section{Conclusion}
\label{sec:conclusion}

We introduce QTestBench, a framework for evaluating LLM-generated quantum test
suites against independent benchmark oracles. It unifies PennyLane QHack and
Qiskit HumanEval through a two-stage pipeline (single-pass generation of
syntactic, semantic, and behavioral pytest modules, followed by validation under
three cohort configurations) and reports executability, correctness acceptance,
bug-detection rejection, test quality, checker-input coverage, and composite
quality with consistent definitions across SDKs.

Experiments with Claude Opus~4.6, DeepSeek
V3.2, Gemini~3 Pro,
GPT-5.4, and Qwen3-Coder show
that runnable quantum tests are achievable ($\mathrm{ES}\approx1.00$ on QHack for
four models), but discrimination remains the primary bottleneck: leading
generators reach only moderate bug detection (Claude Opus~4.6,
$\mathrm{TQS}=0.71$ on QHack), and Qiskit scores are lower and more
task-heterogeneous. By scoring tests against benchmark oracles, QTestBench addresses a dimension of quantum LLM evaluation not covered by solution-centric benchmarks.
\small{
\section*{Acknowledgment}
 This work was supported in part by the NYUAD Center for Quantum and Topological Systems (CQTS), funded by Tamkeen under the NYUAD Research Institute grant CG008, and the Center for Cyber Security (CCS), funded by Tamkeen under the NYUAD Research Institute Award G1104.}
\small{\section*{Code availability} The source code and experiment configurations are available at \url{https://github.com/eBrain4Everyone/QTestBench}.}
\bibliographystyle{IEEEtran}
{\setstretch{0.88}

{\scriptsize
\bibliography{refs}}}

\end{document}